\pdfoutput=1
\documentclass[11pt]{article}

\usepackage{ACL2023}

\usepackage{times}
\usepackage{latexsym}

\usepackage{booktabs}       % professional-quality tables
\usepackage{multirow}
\usepackage{threeparttable}
\usepackage{graphicx}

\usepackage[T1]{fontenc}
\usepackage[utf8]{inputenc}

\usepackage{microtype}

\usepackage{inconsolata}

\usepackage{subfigure}
\usepackage{xspace}

\def\figref#1{Figure~\ref{fig:#1}}
\def\figlabel#1{\label{fig:#1}\label{p:#1}}

\def\tabref#1{Table~\ref{tab:#1}}

\def\tablabel#1{\label{tab:#1}\label{p:#1}}

\def\secref#1{\S\ref{sec:#1}}
\def\seclabel#1{\label{sec:#1}}
\def\qref#1{Eq.~\ref{eqn:#1}}

\def\eqlabel#1{\label{eqn:#1}}

\def\method{\textsc{ProFiT}\xspace}

\newcommand\blfootnote[1]{%
  \begingroup
  \renewcommand\thefootnote{}\footnote{#1}%
  \addtocounter{footnote}{-1}%
  \endgroup
}

\title{Is Prompt-Based Finetuning Always Better than Vanilla Finetuning? \\Insights from Cross-Lingual Language Understanding}

\author{
Bolei Ma$^{\star}$ $^1$~~~
Ercong Nie$^{\star}$ $^1$$^,$$^2$~~~
Helmut Schmid$^1$~~~
Hinrich Schütze$^1$$^,$$^2$~~~
\\
$^1$Center for Information and Language Processing (CIS), LMU Munich, Germany
\\
$^2$Munich Center for Machine Learning (MCML), Munich, Germany \\
\texttt{bolei.ma@campus.lmu.de ~~~ nie@cis.lmu.de}
}

\begin{document}
\maketitle
\begin{abstract}
%Prompt-based learning has emerged as a promising approach. 
Multilingual pretrained language models (MPLMs) have demonstrated substantial performance improvements in zero-shot cross-lingual transfer across various natural language understanding tasks by finetuning MPLMs on task-specific labelled data of a source language (e.g. English) and evaluating on a wide range of target languages. Recent studies show that prompt-based finetuning surpasses regular finetuning in few-shot scenarios. However, the exploration of prompt-based learning in multilingual tasks remains limited. In this study, we propose the \textbf{\method} pipeline to investigate the cross-lingual capabilities of \textbf{Pro}mpt-based \textbf{Fi}ne\textbf{t}uning. We conduct comprehensive experiments on diverse cross-lingual language understanding tasks (sentiment classification, paraphrase identification, and natural language inference) and empirically analyze the variation trends of prompt-based finetuning performance in cross-lingual transfer across different few-shot and full-data settings. Our results reveal the effectiveness and versatility of prompt-based finetuning in cross-lingual language understanding. Our findings indicate that prompt-based finetuning outperforms vanilla finetuning in full-data scenarios and exhibits greater advantages in few-shot scenarios, with different performance patterns dependent on task types. Additionally, we analyze underlying factors such as language similarity and pretraining data size that impact the cross-lingual performance of prompt-based finetuning. Overall, our work provides valuable insights into the cross-lingual prowess of prompt-based finetuning. 
%The codes for our work are available here: \url{https://github.com/boleima/ProFiT}.
%~\citep{gao-etal-2021-making, liu2023pre}
\blfootnote{$^\star$ Equal Contribution.}
\end{abstract}

\begin{figure*}[ht]
	\centering  
	\subfigure[Vanilla finetuning]{  
		\begin{minipage}{.485\linewidth}
			\centering    
			\includegraphics[width=.96\linewidth]{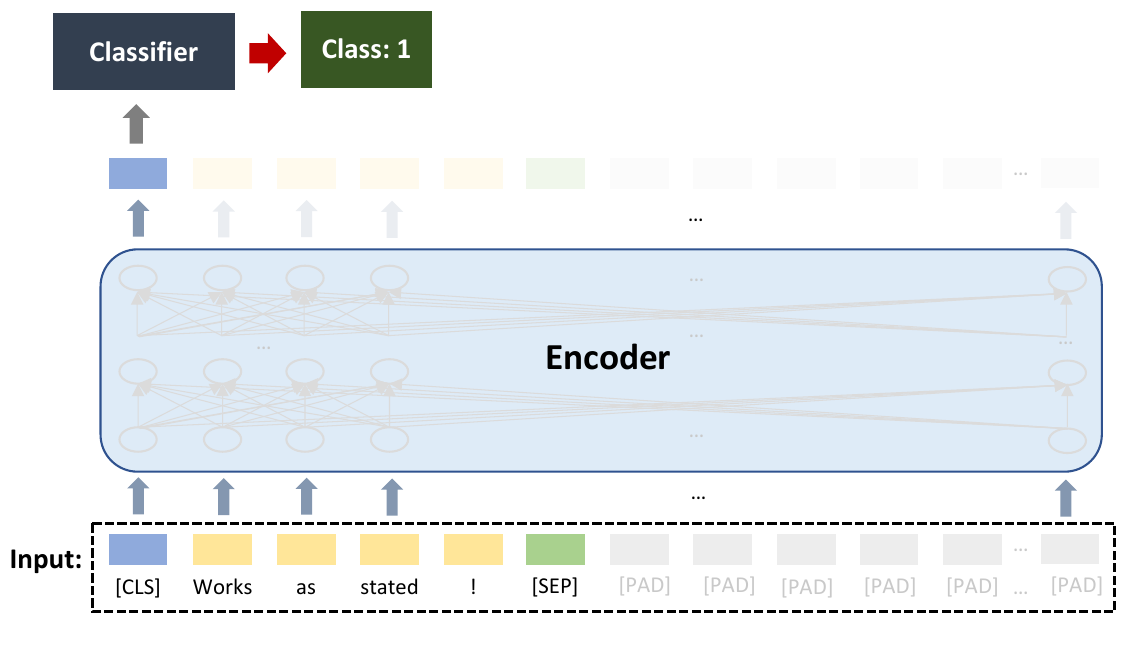} 
		\end{minipage}
	}
        \subfigure[Prompt-based finetuning]{ 
		\begin{minipage}{.485\linewidth}
			\centering    
			\includegraphics[width=.96\linewidth]{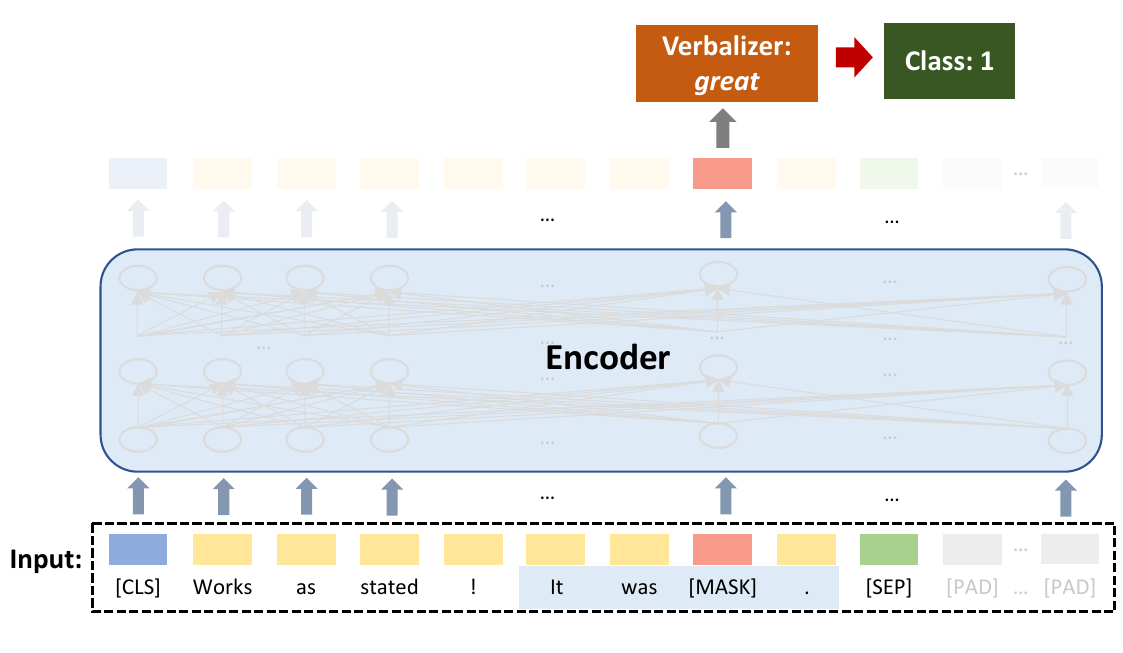}
                \figlabel{prompt_ft}
		\end{minipage}
	}
\caption{The comparion of vanilla finetuning and prompt-based finetuning. [CLS], [SEP], [MASK], [PAD] are special tokens in the encoder vocabulary. The verbalizer is a function mapping from the task label set to a subset of the encoder vocabulary. Input tokens in blue represent the prompt pattern.}    
    \figlabel{comparison}   

\end{figure*}

\section{Introduction}

Pretrained language models (PLMs)~\citep{devlin-etal-2019-bert, yang2019xlnet, radford2019language}, trained on massive amounts of unlabelled data in a self-supervised manner, have shown strong performance after finetuning on task-specific labelled data for a given downstream task, such as sentence classification~\citep{zhuang-etal-2021-robustly}, text summarization~\citep{zhang2020pegasus}, or dialogue generation~\cite{liu-etal-2023-pvgru}.
\textit{Prompt-based learning}~\citep{brown2020language, schick-schutze-2021-exploiting, schick-schutze-2021-shot, schick-schutze-2021-just} has recently emerged as a notable advancement, surpassing regular finetuning approaches in few-shot scenarios~\cite{liu2023pre}. 
In prompt-based learning, downstream tasks are reformulated to resemble the types of problems tackled during the PLM's original pretraining by using a textual prompt. 
For example, in~\figref{prompt_ft}, an input sentence of the binary sentiment analysis task ``\texttt{Works as stated!}'' can be reformulated with a prompt pattern $P(X) = X \circ$ ``\texttt{It was [MASK]}.'' as ``\texttt{Works as stated! It was [MASK].}'' where $\circ$ is the string concatenation operator.
We use a \textit{verbalizer} which maps the class label to a \textit{label word}. In this example, the class labels \textsc{positive} and \textsc{negative} can be verbalized as ``\texttt{great}'' and ``\texttt{bad}''. By comparing the probabilities of the label words ``\texttt{great}'' and ``\texttt{bad}'' as fillers of the \texttt{[MASK]} token, we can predict the correct class label. 
In the example above, a natural language understanding (NLU) task is transformed into a masked language modeling (MLM) problem, which is the same as the PLM's pretraining objective.

The reformulated input can be used for finetuning, i.e.\ \textit{prompt-based finetuning}. \figref{comparison} shows the difference between prompt-based finetuning and vanilla finetuning. Vanilla finetuning solely relies on the hidden embedding of the \texttt{[CLS]} token. In contrast, prompt-based finetuning makes use of both the semantic information from the task labels and the prior knowledge encoded in the pretraining phase. Recent empirical studies of few-shot learning showed advantages of prompt-based finetuning over vanilla finetuning~\citep{gao-etal-2021-making, li-liang-2021-prefix}. 

When applied to multilingual pretrained language models (MPLMs), prompt-based finetuning also enables zero-shot\footnote{In this paper, ``zero-shot'' in ``zero-shot cross-lingual tranfer'' refers to the number of target language training data, i.e., no target language data is provided, while ``few-shot'' in ``few-shot finetuning'' refers to the source language used for finetuning, i.e., a few source language data is provided for the finetuning of the MPLM. The finetuned model is then zero-shot transferred to target language.} cross-lingual transfer. MPLMs such as mBERT~\citep{devlin-etal-2019-bert} and XLM-R~\citep{conneau-etal-2020-unsupervised} are pretrained on huge multilingual corpora and show strong multilinguality~\citep{pires-etal-2019-multilingual, dufter-schutze-2020-identifying, liang2021locating}. They have become the dominant paradigm for zero-shot cross-lingual transfer, where annotated training data is available for some source language (e.g.\ English) but not for the target language~\citep{wu-dredze-2019-beto, pmlr-v119-hu20b}. 
\citet{zhao-schutze-2021-discrete}
proposed % made initial explorations in introducing
prompt-based finetuning
for %into
cross-lingual transfer. Their work focused on few-shot finetuning. Their experimental results for the natural language inference task showed that prompt-based finetuning performed better in few-shot cross-lingual transfer than vanilla finetuning. 
However, prior studies failed to examine whether prompt-based learning is also advantageous
when training data is not scarce. %in the full-data setup. 
Therefore, we conduct a comprehensive investigation on diverse cross-lingual language understanding tasks in both full-data and few-shot settings in order to shed more light on the cross-lingual capabilities of prompt-based finetuning.

In contrast to most previous research on prompting, our work is not restricted to monolingual or few-shot scenarios. Instead we explore a wide range of few-shot settings. We adopt a multilingual perspective and aim to uncover the nuances of performance variations associated with prompt-based finetuning. To this end, we implement the \method pipeline and carry out an extensive set of experiments encompassing three representative cross-lingual language understanding tasks: 
% a multi-class sentiment classification task (Amazon Reviews) which represents single-sentence classification, and two sentence pair classification tasks including a binary paraphrase identification task (PAWS-X) and a multi-class natural language inference task (XNLI). All three tasks represent different aspects of NLU. 
sentiment analysis (Amazon Reviews), paragraph identification (PAWS-X), and natural language inference (XNLI). Our task selection covers single-sentence classification, sentence pair classification and inference task, considering both binary and multi-fold classifications.
Our work provides insights into the effectiveness and versatility of prompt-based finetuning in cross-lingual language understanding.

\paragraph{Research Questions and Contributions.} In this work, we analyze how the performance of prompt-based finetuning varies with the size of the labelled source language data for zero-shot cross-lingual transfer tasks. We examine a wide range of factors which could have an impact on cross-lingual transfer performance. We attempt to address the following pivotal research questions:
\subparagraph{RQ1}\textit{Does prompt-based finetuning outperform vanilla finetuning in the full-data scenario in different NLU tasks?}

We propose the \method pipeline for systematically conducting the cross-lingual transfer experiments. We carry out zero-shot cross-lingual transfer experiments on three different NLU tasks using all the available English training data. 
By comparing the results of vanilla finetuning and \method for different MPLMs, we find that in the full-data scenario, \method still achieves better cross-lingual performance than vanilla finetuning.

\subparagraph{RQ2}\textit{Is prompt-based finetuning always better than vanilla finetuning?}

We investigate how the cross-lingual performance depends on the size of the English training data. Our findings substantiate that the \method exhibits greater advantages in few-shot scenarios compared to full-data scenarios. The specific patterns of performance change are contingent upon the task types.

\subparagraph{RQ3}\textit{What underlying factors could affect the cross-lingual performance of \method?}

We extensively analyze the factors that could influence the cross-lingual performance of \method, encompassing language similarity, pretraining data size of target languages, etc.

\section{Related Work}
\paragraph{Prompt-Based Learning}
GPT-3~\cite{brown2020language} has sparked research in
prompt-based methods. Recent advances %in prompt-based learning
include automatic generation of prompt verbalizers and
patterns~\citep{schick-etal-2020-automatically,
  shin-etal-2020-autoprompt}, soft
prompting~\citep{qin-eisner-2021-learning}, prefix
tuning~\citep{li-liang-2021-prefix}, P-tuning~\citep{liu-etal-2022-p}, and
retrieval-augmented prompting~\citep{liu2022semantic}.
Most of these methods
focus %focuses
on monolingual scenarios, leaving the cross-lingual capabilities of prompt-based methods largely unexplored.

\paragraph{MPLMs and Zero-Shot Cross-Lingual Transfer}
The advances of MPLMs have positioned them as the standard approach for cross-lingual transfer. MPLMs usually adopt the architecture of some monolingual Transformer-based language model \citep{vaswani2017attention} and are jointly pretrained on large unlabelled multilingual data. For instance, mBERT~\citep{devlin-etal-2019-bert} is based on BERT; XLM-R~\citep{zhuang-etal-2021-robustly} and Glot500-m~\citep{imanigooghari-etal-2023-glot500} are based on  RoBERTa~\citep{conneau-etal-2020-unsupervised}.
A multitude of studies have validated the robust multilinguality exhibited by MPLMs, either through probing the MPLMs themselves~\citep{pires-etal-2019-multilingual} or by identifying the key factors that contribute to their impressive multilinguality~\citep{dufter-schutze-2020-identifying}. 
Recent empirical studies have further demonstrated the remarkable cross-lingual capabilities of MPLMs by finetuning MPLMs on English training sets and then predicting on test sets of other languages~\citep{karthikeyan2020cross, turc2021revisiting}. 
Several benchmarks have been proposed to evaluate the performance of multilingual encoders, including XTREME~\citep{hu2020xtreme}, XTREME-R~\citep{ruder-etal-2021-xtreme}, Taxi1500~\citep{ma2023taxi1500} and XGLUE~\citep{liang-etal-2020-xglue}.

\paragraph{Multilingual Prompt Learning}
While prompting has proven successful in English, the application of prompting techniques in multilingual tasks has yet to be thoroughly explored and extensively studied.
\citet{zhao-schutze-2021-discrete} first investigated prompt-based methods for cross-lingual transfer with different prompt forms and verbalizers. 
Recent follow-up studies introduced mask token augmentation~\citep{zhou2022enhancing} and unified multilingual prompts~\citep{huang2022zero} for zero-shot cross-lingual transfer.
Despite the growing attention garnered by these methods in the context of few-shot scenarios across various NLP tasks, there remains a dearth of comprehensive investigations into the variations of prompt-based learning methods across different few-shot settings and full-data settings. 
\citet{tu-etal-2022-prompt} focused on an alternative prompting approach for cross-lingual transfer in full-data scenarios. In contrast to prompt-based finetuning, they introduced additional prompt parameters to PLMs and exclusively updated these parameters during the finetuning process.
A more recent work~\citep{shi2023dont} combined prompt-based finetuning and continued pretraining, but it was limited to monolingual scenarios.

In contrast to the aforementioned previous studies, our work provides a comprehensive investigation of prompt-based finetuning for cross-lingual transfer in both few-shot and full-data scenarios. Furthermore, we empirically analyze the variation of prompt-based finetuning performance across different few-shot settings.

\section{Methodology}
The purpose of this study is to improve the cross-lingual transfer performance of vanilla finetuning. In vanilla settings of zero-shot cross-lingual transfer, the MPLM is directly finetuned with training data in a source language (English). The finetuned model is then applied to predict the test data in target languages. 

\begin{figure*}
\centering
\includegraphics[scale=0.65]{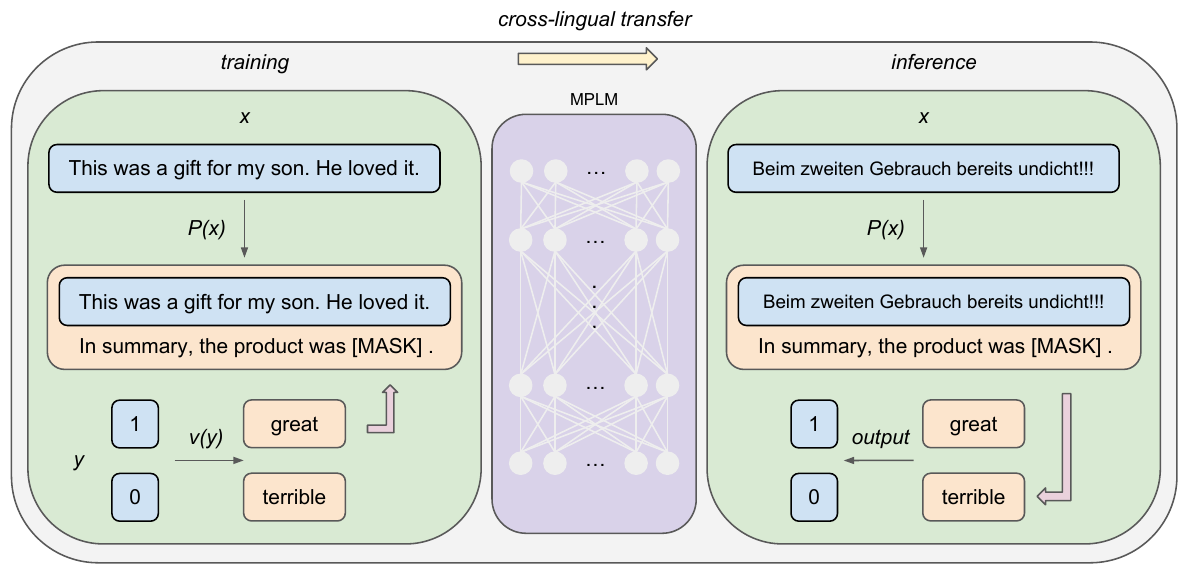}
\caption{\method pipeline of training and cross-lingual transfer with examples. $X$ is an input sentence and $P(X)$ denotes the prompt pattern which reformulates the input into a prompt. $v(y)$ is the verbalizer which maps each class label $y$ onto a word from the source language vocabulary.}
\figlabel{profit}
\end{figure*}

In prompt-based learning, we need a pattern-verbalizer pair (PVP) \citep{schick-schutze-2021-exploiting} consisting of (i) a \emph{prompt pattern} which converts the input text into a cloze-style question with a mask token, and (ii) a representative word (called \emph{verbalizer}) for each possible class. In our \method approach, a PVP is combined with training data in English during finetuning. As the \emph{training} block in \figref{profit} shows, a prompt pattern such as $P(X)=X \circ$ ``\texttt{In summary, the product was [MASK].}'' is filled with an input example $X$ ``\texttt{This was a gift for my son. He loved it.}'' A verbalizer such as \{0 $\rightarrow$ ``terrible'', 1 $\rightarrow$ ``great''\} is used to map the original labels \{0,1\} onto words. The MPLM takes the filled pattern ``\texttt{This was a gift for my son. He loved it. In summary, the product was [MASK].}'', as input and returns for each of the two verbalizers ``terrible'' and ``great'' its probability of being the masked token. Thus, it uses the PVP to reformulate the sentence classification task of vanilla finetuning into a masked token prediction task. 

More formally, let $D$=$\{(X_1, y_1), ... , (X_n, y_n)\}$ denote the set of training examples in the source language, where $X_1,...,X_n$ are text samples and $y_1,...,y_n$ are class labels from a label set $Y$. The prompt pattern $P(.)$ transforms an input sentence $X$ into a cloze-style question with a masked token. The pretrained language model $M$ with trainable parameters $\theta$ performs masked token prediction and returns the probabilities $p = M(P(X),\theta)$ of all candidate words for the masked token in $P(X)$. The verbalizer $v(.)$ is a bijective mapping from the set of class labels $Y$ to a set of verbalised words $V$ from the source language vocabulary. We predict the class $\hat{y}$ whose verbalizer $v(\hat{y})$ received the highest probability from model $M$:
\begin{equation}
  \eqlabel{argmax}
\hat{y} = \arg\max_{y\in Y} p(v(y))
\end{equation} 
We finetune the parameters $\theta$ of model $M$ by minimizing the cross-entropy loss function $\ell$ on D: % cross-entropy loss: confirmed!
\begin{equation}
  \hat{\theta} = \arg\max_\theta \sum_{(X,y) \in D} \ell(v(y), M(P(X), \theta))
\end{equation} 
The model with the finetuned parameters $\hat{\theta}$ is used to predict the class labels of the target language examples $D^{\prime} = \{X_1^{\prime}, ... , X_n^{\prime}\}$ using the same prompt pattern and verbalizer as during finetuning (see \emph{inference} block in \figref{profit}). The best label $y_i'$ for each example $X_i'$ is predicted according to \qref{argmax}.

In contrast to vanilla finetuning, prompt-based methods such as
\method only transform the training data with the prompt pattern $P$
and the verbalizer $v$, but leave the model architecture unchanged.
thus not hindering the efficiency of Vanilla much \citep{shi2023dont}.  No extra
parameters have to be trained from scratch. By
reformulating the sentence classification task into a masked token
prediction (MTP) task, we can better take advantage of the knowledge that
the model has acquired during MTP pretraining.
% we deleted the last sentence, without providing much information
% \red{I don't really understand the following argument:} incorporating
% the prompt during the finetuning process, greater emphasis is placed
% on the prompt itself, thereby enhancing overall performance.

In the cross-lingual setting, we simply apply the same functions $P$
and $v$ to the target language examples without further modifications.

\section{Experimental Setups}
\subsection{Datasets}
In order to investigate the performance on diverse NLU tasks, three representative different classification tasks on NLU are selected for evaluation in this work: sentiment analysis on Amazon product reviews \citep{keung-etal-2020-multilingual}, paraphrase identification on PAWS-X \citep{yang-etal-2019-paws}, and natural language inference on XNLI \citep{conneau-etal-2018-xnli}.

\textbf{Amazon Reviews Dataset} \citep{keung-etal-2020-multilingual} contains product reviews with 5 star ratings from 1 to 5. The multilingual version of this dataset consists of test data in English and 5 other languages. We use the following prompt pattern $P(X)$ and verbalizer $v(y)$ for each review example $(X, y)$:
\begin{itemize}
\item $P(X) = X \circ$ ``All in all, it was [MASK].''
\item $v(1)= $ ``terrible'', $v(2) = $ ``bad'', \newline $v(3) = $ ``ok'', $v(4) = $``good'', $v(5) = $ ``great''
\end{itemize}

\textbf{PAWS-X} is a multilingual version of PAWS \citep{zhang2019paws}, which consists of challenging paraphrase identification pairs from Wikipedia and Quora. Each data item comprises two sentences. The task is to predict whether the two sentences are paraphrases. The labels are binary: 1 for paraphrase, 0 for non-paraphrase. PAWS-X consists of datasets in English and 6 other languages. For a given sentence pair $X_1$ and $X_2$, we design the pattern and verbalizer as:
\begin{itemize}
\item $P(X_1, X_2) = X_1 \circ$``? [MASK],~'' $\circ X_2$
\item $v(0) = $ ``Wrong'', $v(1) =$ ``Right''
\end{itemize}

\textbf{XNLI} is a multilingual version of the MultiNLI dataset \citep{williams-etal-2018-broad}. The text in each data item consists of two sentences. Sentence A is the premise and sentence B is the hypothesis. The task is to predict the type of inference between the given premise and hypothesis among the three types: ``entailment'' (0), ``neutral'' (1), and ``contradiction'' (2). It is a kind of multi-class natural language inference task. XNLI consists of datasets in English and 14 other languages. For a given sentence pair $X_1$ and $X_2$, we design the pattern and verbalizer as:
\begin{itemize}
\item $P(X_1, X_2) = X_1 \circ$ ``? [MASK],~'' $\circ X_2$
\item $v(0) =$ ``Yes'', $v(1) =$ ``Maybe'', $v(2) = $ ``No''
\end{itemize}

\subsection{Baseline}

The following baselines are considered and compared to our \method approach:

\textbf{MAJ} The majority baseline. It always assigns the majority class from the training data.

\textbf{Direct} The pattern filled with the input sample is directly fed to the MPLM for prediction, without finetuning. This is the zero-shot scenario.

\textbf{Vanilla} The standard finetuning method which predicts the class from the hidden embedding of the \texttt{[CLS]} token without using a prompt pattern. We use the cross-entropy loss as the objective function for finetuning and AdamW for optimization with a learning rate of 1e-5 and 5 training epochs. The finetuned models are then used to predict the test data.

\subsection{Multilingual Models}
In order to solve the classification tasks with cross-lingual transfer, we use the pretrained multilingual BERT model \citep{devlin-etal-2019-bert} ``\texttt{bert-base-multilingual-cased}'' (M) and the XLM-R model \citep{conneau-etal-2020-unsupervised} ``\texttt{xlm-roberta-base}'' (X) from the Huggingface Transformers library \citep{wolf-etal-2020-transformers}. Both models are evaluated with the methods Vanilla and \method. We repeat all our experiments 5 times with different random seeds. The details about model training and hyperparameter settings can be found  in Appendix \secref{training_details}.

\section{Results}

\subsection{Main Results}
\seclabel{main}

\begin{table}[h]
\renewcommand\arraystretch{1.3}
\setlength\tabcolsep{3pt}
\centering
\footnotesize
\begin{tabular}{lcccc}
\toprule
 & Amazon & PAWS-X & XNLI & Avg. \\
 \midrule
 MAJ & 20 & 55.81 &33.33& 36.17 \\
 \midrule
 Direct-mBERT &20.21& 45.05& 35.05&33.44 \\
 Vanilla-mBERT & 42.97& 80.24& 65.05 & 62.75\\
 \method-mBERT & \textbf{43.98} & \textbf{82.16}& \textbf{65.79} &\textbf{63.98}\\
 \midrule
 Direct-XLM-R & 21.98 & 51.10& 35.68 & 36.25\\
 Vanilla-XLM-R & 54.56 &82.51&73.61 &70.22 \\
 \method-XLM-R & \textbf{54.66} &\textbf{82.73} &\textbf{73.82} &\textbf{70.40}\\ 
\bottomrule
\end{tabular}
\caption{Overview of results}
\tablabel{overview}
\end{table}

\begin{table*}[h]
\renewcommand\arraystretch{1.2}
\setlength\tabcolsep{2.5pt}
\centering
\scriptsize
\begin{tabular}{lccccccccccccccccccc}
\toprule
 Task & Model&en &ar&	bg	&de&el	&es	&fr	&hi	&ja &ko &ru	&sw	&th	&tr &ur	&vi	&zh	&avg. \\
 \midrule
 \multirow{4}*{Amazon}& Vanilla-M & 58.92 &- &-&45.69	&-&	48.02&	47.45	&-&35.07	&-&-&-&-&-&-&-&\textbf{38.63}	&42.97 \\
 &\method-M& \textbf{59.05} &- &-&\textbf{46.66}	&-&	\textbf{49.30}&	\textbf{48.38}	&-&\textbf{37.31}	&-&-&-&-&-&-&-&38.26	&\textbf{43.98} \\
 \cmidrule(l){2-20}
 & Vanilla-X& 59.61 &- &-&\textbf{60.14}	&-&	55.24&	55.66	&-&51.93	&-&-&-&-&-&-&-&49.82	&54.56 \\
 &\method-X& \textbf{60.06} &- &-&59.60	&-&	\textbf{55.72}&	\textbf{55.89}	&-&\textbf{52.34}	&-&-&-&-&-&-&-&\textbf{49.75}	&\textbf{54.66} \\
\midrule
 \multirow{4}*{PAWS-X}& Vanilla-M& 93.85 &- &-&84.94	&-&	87.11&	86.55	&-&73.39	&72.44&-&-&-&-&-&-&77.01	&80.24\\
 &\method-M& \textbf{94.21} &- &-&\textbf{86.06}	&-&	\textbf{88.17}	&\textbf{87.91}	&-&\textbf{75.79}	&\textbf{75.82}&-&-&-&-&-&-&\textbf{79.22}	&\textbf{82.16} \\
 \cmidrule(l){2-20}
 & Vanilla-X& 94.33 &- &-&86.92	&-&	88.55&	89.04	&-&\textbf{76.07}	&74.71&-&-&-&-&-&-&79.75	&82.51\\
 &\method-X& \textbf{94.90} &- &-&\textbf{87.06}	&-&	\textbf{88.87}	&\textbf{88.86}	&-&75.53	&\textbf{75.40}&-&-&-&-&-&-&\textbf{80.63}	&\textbf{82.73} \\
 \midrule
 \multirow{4}*{XNLI}& Vanilla-M& 82.57 &65.12 & 68.97 & 71.40 & 66.30  & 74.22 & 73.68 & 60.02 &-&-& 68.95 & 50.24 & 53.15 & 62.02 & 57.96 & 69.80 & 68.91 & 65.05 \\
 &\method-M& 82.57 &\textbf{65.55} & \textbf{69.47} & \textbf{71.57} & \textbf{67.43} & \textbf{75.10} & \textbf{74.57} & \textbf{60.57} &-&-& \textbf{69.55} & \textbf{51.13} & \textbf{54.58} & \textbf{62.64} & \textbf{58.04} & \textbf{70.74} & \textbf{70.08} & \textbf{65.79}  \\
 \cmidrule(l){2-20}
 & Vanilla-X& 84.91 &\textbf{71.86} & 77.78 & 76.86 & 75.96 & 79.25 & 78.21 & 69.92 &-&-& \textbf{75.79} & \textbf{65.21} & 72.02 & 73.12 & 66.07 & 74.71 & 73.72 & 73.61  \\
 &\method-X& \textbf{84.97} &71.81 & \textbf{77.92} & \textbf{77.35} & \textbf{76.11} & \textbf{79.31} & \textbf{78.75} & \textbf{70.10} &-&-& 75.43 & 65.13 & \textbf{72.39} & \textbf{73.23} & \textbf{66.95} & \textbf{75.05} & \textbf{73.92} & \textbf{73.82}  \\
\bottomrule
\end{tabular}
\caption{Detailed cross-lingual performance results on three classification tasks. When calculating the average (avg.), due to the aim of zero-shot cross-lingual transfer, the performance results of the source language English are not taken into account. Model M stands for mBERT, and X for XLM-R.}
\tablabel{overview-full}
\end{table*}

\tabref{overview} gives an overview of the experimental results. \method outperforms the MAJ baseline with both mBERT and XLM-R for all three classification tasks. \method also outperforms the Direct and Vanilla baselines in both mBERT and XLM-R settings: When trained with mBERT, the performance is improved by \textbf{23.77\%}, \textbf{37.11\%} and \textbf{30.74\%} compared to Direct on Amazon, PAWS-X and XNLI respectively, and by \textbf{1.01\%}, \textbf{1.92\%} and \textbf{0.74\%} compared to Vanilla. When trained with XLM-R, the performance is improved by \textbf{32.68\%}, \textbf{31.63\%} and \textbf{38.14\%} compared to Direct, and by \textbf{0.10\%}, \textbf{0.22\%} and \textbf{0.21\%} compared to Vanilla respectively.

While \method outperforms all baselines on all three tasks, the degree of improvement differs. The improvements of \method over Vanilla when trained with mBERT (\textbf{+1.23\%}) are larger than the improvements when trained with XLM-R (\textbf{+0.18\%}). 

We further conducted T-tests for results of Vanilla and \method with different random seeds (see \secref{training_details} for the seeds). \tabref{ttest} shows the T-test results with $p$ values for each task with mBERT and XLM-R models. We can see that the $p$ values of all three tasks with mBERT model are under 0.05, indicating that the performance gain of \method is significant with mBERT, while the $p$ values of all three tasks with XLM-R model are bigger than 0.05, showing no significant performance difference. 

\begin{table}[h]
\renewcommand\arraystretch{1.3}
\setlength\tabcolsep{6pt}
\centering
\footnotesize
\begin{tabular}{lccc}
\toprule
Model & Amazon         & PAWS-X         & XNLI            \\
\midrule
mBERT & 0.005 & 0.003 & 0.005  \\
XLM-R & 0.40$^*$   & 0.46$^*$   & 0.44$^*$   \\
\bottomrule
\end{tabular}
\caption{T-Test results ($p$) for results of Vanilla and \method with different random seeds. Insignificant results with a $p$ value $>$ 0.05 are marked with $^*$.}
\tablabel{ttest}
\end{table}

One reason for the performance difference of the two models could be that the XLM-R model was pretrained on far more data than mBERT and is also much bigger, so that the Vanilla performance with XLM-R finetuning is much better than with mBERT in cross-lingual context \citep{conneau-etal-2020-unsupervised, lauscher-etal-2020-zero}, leaving less space for improvement.

%delete this part: On the other hand, a good improvement is found on Amazon and PAWS-X with mBERT, while XNLI only shows a slight improvement. It could be stated that the multi-class inference task is more difficult than the paraphrase task and sentiment analysis. In addition, previous research has indicated that creating cloze-style patterns and exploring the solution domain is difficult for inference tasks \citep{schick-schutze-2021-exploiting, webson-pavlick-2022-prompt, nie-etal-2023-cross}.
%\red{IMO, the results for mBERT and XLM-R together do not support the claim that XNLI is a more difficult task than Amazon and PAWS-X: First of all, the absolute performance of all systems is much better on XNLI than on Amazon. (The 5 classes might make Amazon a hard problem.)  Second, the improvement of \method over Vanilla for PAWS-X with mBERT is closer to XNLI than to Amazon. The improvement of \method over Vanilla for XLM-R is similar for PAWS-X and XNLI and larger than for Amazon despite the fact that Amazon leaves more room for improvement.}

A detailed overview of the cross-lingual performance of \method compared to Vanilla for each target language is presented in \tabref{overview-full}. Although the overall performance of \method is better than Vanilla for all three tasks in both mBERT and XLM-R settings, individual differences between languages can be noticed. On Amazon, with mBERT, the improvement in Japanese (ja) (\textbf{+2.24\%}) is far greater than on average, whereas Chinese (zh) shows no improvement (\textbf{-0.37\%}); with XLM-R, \method performs slightly worse than Vanilla on both Chinese with \textbf{-0.07\%} and German (de) with \textbf{-0.54\%}. On PAWS-X, Korean (ko) shows a larger improvement (\textbf{+3.38\%}) than average with mBERT, and with XLM-R, whereas French (fr) (\textbf{-0.18\%}) and Japanese (\textbf{-0.54\%}) show a slightly worse performance than Vanilla. On XNLI, we find improvements for all languages with mBERT, and with XLM-R, Arabic (ar) (\textbf{-0.06\%}), Russian (ru) (\textbf{-0.36\%}), and Swahili (sw) (\textbf{-0.08\%}) show slightly worse performance than Vanilla. 

We conclude that the performance gain of \method over Vanilla depends on the models and languages. In \secref{cross}, we will further investigate how linguistic factors influence cross-lingual transfer performance.

\subsection{Few-shot Ablations}

Previous studies show that the prompt framework is more effective than finetuning when training data is scarce \citep{zhao-schutze-2021-discrete, qi-etal-2022-enhancing}. We investigated how the performance changes as the number of training samples $K$ increases in few-shot settings. The training and validation data are randomly sampled with $K \in \{1, 2, 4, 8, 16, 32, 64, 128, 256, 512, 1024\}$ shots per class from the English training data.

\begin{figure}
\centering
\includegraphics[scale=0.42]{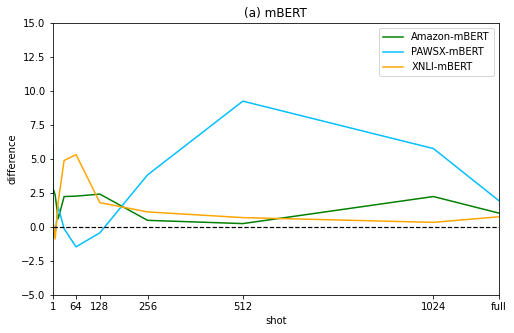}
\includegraphics[scale=0.42]{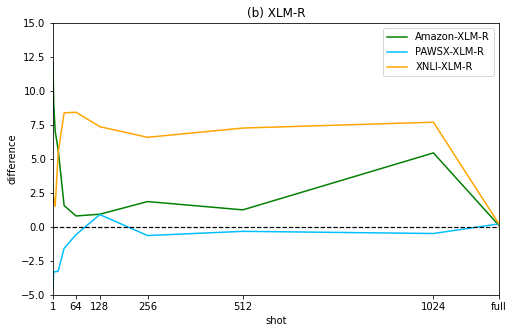}
\caption{Performance difference between \method and Vanilla in different few-shot settings and full training setting on three NLU tasks with both mBERT and XLM-R models.}
\figlabel{dif}
\end{figure}

%\red{Did you interpolate the curves in Figure 3 with splines or so? This is confusing and it is hard to verify your claims from these curves. It would be better to use straight lines between the points. Also, don't label the last position on the x-axis with +inf. Better use ``all data'' or so.
%In the discussion of the results, it would be preferable to concentrate on those results which allow you to draw valid conclusions. You should not try to recap all results from the figure.}

The detailed results of few-shot ablations can be found in \tabref{fewshot_amazon}, \tabref{fewshot_pawsx} and \tabref{fewshot_xnli} in Appendix \secref{detailed_results}. \figref{dif} shows the performance changes on all three tasks with both mBERT and XLM-R models. On the Amazon task, the performance improvement for smaller numbers of shots is greater than for full training. As the number of shots increases, the improvement decreases accordingly. This implies that on the sentiment analysis task, \method is most valuable with small training data. On XNLI, the improvement of \method over Vanilla is first small with in small shots. It then gets greater, as $K$ increases, and drops again, as bigger $K$ towards full data size shows up. We conclude that on NLI tasks such as XNLI, \method is most effective in few-shot settings with a certain number of $K$. On PAWS-X, no obvious difference in few-shot settings can be found with mBERT in small shots, but in bigger shots there is greater improvement with $K \in \{256, 512, 1024\}$; however, with XLM-R, \method shows almost no performance improvement over Vanilla.

%is largest for $K = 64$. The other performance differences are not much different from full-shot. With XLM-R, the performance improvements are larger for all few-shot variants than for full-shot, and among all numbers of shots, the largest improvements are found for $K \in [64,1024] $. We conclude that on NLI tasks such as XNLI, \method is most effective in few-shot settings. 

Overall, sentiment analysis exhibits a clearer performance improvement for smaller numbers of shots, whereas the language inference and paraphrase tasks show greater performance enhancements in few-shot scenarios with larger $K$. This might be due to difficulties with pairwise inputs in these tasks, where we aim to identify the relationship between a pair of sentences. When it comes to transferring knowledge of sentence relationships, more examples are needed for successful learning than in sentiment analysis tasks where semantic information from comparable cross-lingual sentences can be directly transferred.

\section{Cross-Lingual Analysis}
\seclabel{cross}

In previous empirical studies of cross-lingual transfer learning~\citep{lauscher-etal-2020-zero, nie-etal-2023-cross}, several key factors were identified to exert great effect on the cross-lingual performance, including (1) the size of the pretraining corpus for the target language and (2) the similarity between the source and target languages. We analyze how these two factors influence \method's effectiveness for the languages on three tasks. 

The pretraining corpus size of the target languages can be simply measured by the $log_2$ of the number of articles in Wikipedia\footnote{\url{https://meta.wikimedia.org/wiki/List\_of\_Wikipedias}}.

For measuring the similarity between languages, we employ methods from recent studies of language representations. 
In these studies, languages are encoded as vectors according to their various linguistic and typological features. 
With these language vectors, a range of distance metrics, such as Euclidean distance and cosine similarity, can be used to measure the similarity between languages.
\citet{littell-etal-2017-uriel} proposed \textsc{lang2vec} which encodes languages using 5 vectors, with each vector representing a specific language feature.
\citet{oestling2023language} measured the lexical similarity by calculating language vectors based on the ASJP word list database~\citep{wichmann2022asjp}. 
\citet{liu-etal-2023-crosslingual} recently proposed a novel language similarity metric from the perspective of conceptualization across multiple languages.
% To investigate the effects on cross-lingual transfer performance, following \citet{lauscher-etal-2020-zero}, \citet{nie-etal-2023-cross} identified three key factors that greatly impact cross-lingual transfer performance: (1) the size of the pretraining corpus for the source language, (2) the size of the pretraining corpus for the target language, and (3) the similarity between the source and target languages. As the source language of this study is sorely English, we pay therefore more attention on the target language size and the linguistic similarity between English and the target languages. 
In our work, we compute two similarity metrics: (i) a comprehensive linguistic similarity metric based on \textsc{lang2vec}~\citep{littell-etal-2017-uriel} and (ii) a lexical similarity metric based on the ASJP word list database~\citep{oestling2023language}. 

\begin{table*}
\renewcommand\arraystretch{1.5}
\setlength\tabcolsep{3.85pt}
\centering
\scriptsize
\begin{tabular}{l|ccccc|c|cc|c|c|ccccccc}
\toprule
\multirow{2}*{lang} & \multicolumn{5}{r}{Typological \& Phylogenetic Sim. }  & & \multicolumn{2}{r}{Lexical Sim.}  && \multirow{2}*{\textbf{Size}} & \multicolumn{6}{c}{Task Performance}    \\
\cmidrule(){2-10}
\cmidrule(){12-17}
     & SYN & PHO & INV & FAM   & GEO   & \textbf{Sim$_1$}   & \emph{UMAP}               & \emph{SVD}    &\textbf{Sim$_2$}    &           & amazon-M         & amazon-X & pawsx-M & pawsx-X & xnli-M & xnli-X \\
\midrule
ar & 65.47 & 70.06 & 75.88 & 0.00  & 97.04 & \textbf{61.69} & -1.90 & 4.87   & \textbf{1.49}  & \textbf{20.20} & -     & -     & -     & -     & 65.55 & 71.81 \\
bg & 78.78 & 90.45 & 70.02 & 13.61 & 99.01 & \textbf{70.38} & 8.65  & 33.21  & \textbf{20.93} & \textbf{18.15} & -     & -     & -     & -     & 69.47 & 77.92 \\
de & 79.05 & 83.62 & 77.62 & 54.43 & 99.76 & \textbf{78.90} & 83.42 & 76.83  & \textbf{80.13} & \textbf{21.42} & 46.66 & 59.60 & 86.06 & 87.06 & 71.57 & 77.35 \\
el & 73.19 & 95.35 & 64.75 & 14.91 & 98.95 & \textbf{69.43} & 1.24  & 24.81  & \textbf{13.03} & \textbf{17.76} & -     & -     & -     & -     & 67.43 & 76.11 \\
es & 84.97 & 85.81 & 64.99 & 9.62  & 99.59 & \textbf{69.00} & 1.61  & 28.30  & \textbf{14.96} & \textbf{20.83} & 49.30 & 55.72 & 88.17 & 88.87 & 75.10 & 79.31 \\
fr & 76.83 & 75.26 & 73.64 & 9.62  & 99.93 & \textbf{67.06} & 1.34  & 31.76  & \textbf{16.55} & \textbf{21.27} & 48.38 & 55.89 & 87.91 & 88.86 & 74.57 & 78.75 \\
hi & 58.79 & 85.81 & 76.53 & 12.60 & 91.10 & \textbf{64.97} & 1.20  & 21.11  & \textbf{11.16} & \textbf{17.26} & -     & -     & -     & -     & 60.57 & 70.10 \\
ja & 49.63 & 64.44 & 65.92 & 0.00  & 85.65 & \textbf{53.13} & -     & -      & -     & \textbf{20.39} & 37.31 & 52.34 & 75.79 & 75.53 & -     & -     \\
ko & 55.66 & 74.62 & 71.04 & 0.00  & 86.93 & \textbf{57.65} & -0.22 & 12.42  & \textbf{6.10}  & \textbf{19.28} & -     & -     & 75.82 & 75.40 & -     & -     \\
ru & 75.74 & 90.45 & 63.17 & 16.67 & 95.81 & \textbf{68.37} & 8.63  & 32.60  & \textbf{20.62} & \textbf{20.87} & -     & -     & -     & -     & 69.55 & 75.43 \\
sw & 42.26 & 90.91 & 76.16 & 0.00  & 91.50 & \textbf{60.17} & -9.05 & -7.18  & \textbf{-8.12} & \textbf{16.23} & -     & -     & -     & -     & 51.13 & 65.13 \\
th & 65.20 & 81.82 & 78.88 & 0.00  & 85.25 & \textbf{62.23} & -0.21 & 3.82   & \textbf{1.81}  & \textbf{17.25} & -     & -     & -     & -     & 54.58 & 72.39 \\
tr & 43.36 & 85.81 & 68.49 & 0.00  & 98.25 & \textbf{59.18} & -7.80 & -1.56  & \textbf{-4.68} & \textbf{19.00} & -     & -     & -     & -     & 62.64 & 73.23 \\
ur & 50.01 & 0.00  & 71.56 & 12.60 & 92.54 & \textbf{45.34} & 1.35  & 24.92  & \textbf{13.14} & \textbf{17.54} & -     & -     & -     & -     & 58.04 & 66.95 \\
vi & 64.92 & 78.33 & 74.76 & 0.00  & 85.25 & \textbf{60.65} & 0.86  & -18.50 & \textbf{-8.82} & \textbf{20.29} & -     & -     & -     & -     & 70.74 & 75.05 \\
zh & 73.49 & 78.33 & 74.91 & 0.00  & 88.42 & \textbf{63.03} & -     & -      & -     & \textbf{20.37} & 38.26 & 49.75 & 79.22 & 80.63 & 70.08 & 73.92 

  \\
\bottomrule
\end{tabular}
\caption{Overview of language features and task performances with \method for correlation analysis. Language features include typological \& phylogenetic similarities (\textbf{Sim$_1$}), lexical similarities (\textbf{Sim$_2$}), and target language size (\textbf{Size}). Task performance contains the \method results on the three datasets with both mBERT and XLM-R models.}
\tablabel{sim}
\end{table*}

The \textsc{lang2vec} approach provides information-rich vector representations of languages from different linguistic and ethnological perspectives.
%Those vector representations can be used to calculate the similarities between the languages on our tasks. 
We adopt five linguistic categories: syntax (SYN), phonology (PHO), phonological inventory (INV), language family (FAM), and geography (GEO). SYN, PHO and INV are typological categories, and FAM and GEO are phylogenetic categories. 
Given these vectors, we calculate 5 different cosine similarity metrics between English and each target language.

The lexical similarity metric is based on a mean normalized pairwise Levenshtein distance matrix from ASJP. The language vectors used for calculating the lexical similarity are reduced in dimensionality. Two dimensionality reduction methods are employed for calculating the lexical similarity: Uniform Manifold Approximation and Projection (\textit{UMAP})~\citep{mcinnes2018umap} and Singular Value Decomposition (\textit{SVD})~\citep{stewart1993early}.

%As Eq.~\ref{eq:sim} shows, 

The final typological and phylogenetic similarity score \textbf{Sim$_1$} for each language pair is calculated by averaging the 5 similarities of \textsc{lang2vec}. Similarly, the lexical similarity score \textbf{Sim$_2$} is calculated by averaging the similarities of the \textit{UMAP} and \textit{SVD} vectors.
More formally, as Eq.~\ref{eq:sim} shows, let $f$ denote a feature from the feature set $\mathcal{F}_n$ for metric $n$, and let $v_f$ denote the corresponding feature vector. The sim$_1$ and sim$_2$ scores for the source language English (e) and some target language $j$ are then calculated by: 

\begin{equation}  
sim_n (e,j) = \frac{1}{|\mathcal{F}_n|} \sum_{f \in \mathcal{F}_n} \frac {v_f (e) \cdot v_f (j)} {\left\| v_f (e)\right\| _{2}\left\| v_f (j)\right\| _{2}}
\label{eq:sim}
\end{equation}

\tabref{sim} shows a list of language features (typological \& phylogenetic similarities, lexical similarities, and target language size) and task performances with \method for the following correlation analysis. The language similarities, namely the typological \& phylogenetic similarities (\textbf{Sim$_1$}) and lexical similarities (\textbf{Sim$_2$}) refer to the similarity between each language and English, based on the above introduced language vectors. Sim$_1$ and Sim$_2$ are calculated by Eq. \ref{eq:sim}. \emph{ja} and \emph{zh} are not included in \citet{oestling2023language}'s original language sets, thus these two values are missing for the lexical similarities. The target language size (\textbf{Size}) is calculated by the $log_2$ of the number of articles in Wikipedia.

%A detailed overview of the language features and task performance is presented in \tabref{feat_per} in Appendix \secref{detailed_results}.

Based on the obtained language features and experimental results of task performance with \method, we did a correlation analysis. \tabref{cor} shows the results of the two correlation tests on each task. 

According to the results of Pearson and Spearman tests and the $p$ values, the two factors, namely, both the size of pretraining data for the target language and the similarity of typological and phylogenetic features of languages (sim$_1$) have a significant positive correlation with the improvement of cross-lingual performance especially on XNLI, with both \method-M and \method-X models. Only the correlations calculated with the similarity of lexical features (sim$_2$) show some insignificant results. Furthermore, on XNLI, the correlation with language similarity is stronger with \method-X, while the correlation with target data size is stronger with \method-M. We argue that the XLM-R model is bigger than mBERT, so that the linguistic features have more effect on the performance, while for the smaller model mBERT the data size plays a greater role, which further reveals our findings in \secref{main} that the applied pretrained model for finetuning has an impact on the \method performance. 

On PAWS-X and Amazon, we find weak correlations with the proposed factors, which could result from the limitation of languages in test data: XNLI comprises 15 different languages, whereas PAWS-X and Amazon only contain 7 and 6 languages in the test set, respectively. Thus weaker correlations have been found. 

% To sum up, the two factors (language similarity and size) have indeed effects on the cross-lingual performance, and we find significant correlations when the test set contains a bigger amount of languages. 
To sum up, language similarity and size are two factors that impact the cross-lingual performance in our study, and we find significant correlations when the test set contains a larger amount of languages. 

\begin{table}
\renewcommand\arraystretch{1.5}
\setlength\tabcolsep{2.6pt}
\centering
\scriptsize
\begin{tabular}{l|c|c|cc|cc|cc}
\toprule

\multirow{2}*{Task}   & \multirow{2}*{Model} & \multirow{2}*{Stat.} & \multicolumn{2}{c}{sim$_1$}     & \multicolumn{2}{c}{sim$_2$} & \multicolumn{2}{c}{Size}         \\
\cmidrule(){4-9}
        &&& $corr.$ & $p$        & $corr.$ & $p$   & $corr.$ & $p$      \\
\midrule
\multirow{4}*{Amazon} & \multirow{2}*{\method-M} & P & 0.73 & 0.16$^*$     & -0.95 & 0.21$^*$ & 0.81 & 0.09$^*$     \\
       &                          & S & 0.70 & 0.19$^*$     & -1.00 & 0.00 & 0.50 & 0.39$^*$     \\
       \cmidrule(){2-9}
       & \multirow{2}*{\method-X} & P & 0.80 & 0.10$^*$     & 1.00  & 0.01 & 0.92 & 0.03     \\
       &                          & S & 0.80 & 0.10$^*$     & 1.00  & 0.00 & 1.00 & 1e-24 \\
\midrule
\multirow{4}*{PAWS-X} & \multirow{2}*{\method-M} & P & 0.82 & 0.05     & 0.31  & 0.69$^*$ & 0.82 & 0.04     \\
       &                          & S & 0.83 & 0.04     & 0.20  & 0.80$^*$ & 0.60 & 0.21$^*$     \\
       \cmidrule(){2-9}
       & \multirow{2}*{\method-X} & P & 0.83 & 0.04     & 0.34  & 0.66$^*$ & 0.84 & 0.04     \\
       &                          & S & 0.77 & 0.07$^*$     & 0.20  & 0.80$^*$ & 0.71 & 0.11$^*$     \\
\midrule
\multirow{4}*{XNLI}   & \multirow{2}*{\method-M} & P & 0.57 & 0.03     & 0.43  & 0.14$^*$ & 0.86 & 9e-05 \\
       &                          & S & 0.59 & 0.03     & 0.53  & 0.06$^*$ & 0.90 & 1e-05 \\
       \cmidrule(){2-9}
       & \multirow{2}*{\method-X} & P & 0.72 & 4e-03 & 0.43  & 0.14$^*$ & 0.70 & 5e-03 \\
       &                          & S & 0.77 & 1e-03 & 0.63  & 0.02 & 0.72 & 4e-03 \\
\bottomrule
\end{tabular}
\caption{Correlations between task performance and language similarities (sim$_1$ \& sim$_2$) and target language size. P stands for Pearson test and S for Spearman test. Insignificant results with a $p$ value $>$ 0.05 are marked with $^*$.}
\tablabel{cor}
\end{table}

\section{Conclusion}

In our work, we introduce \method for zero-shot cross-lingual transfer, a pipeline which reformulates input examples into cloze-style prompts and applies the input examples with the prompts and its verbalizers as masked token to finetuning, changing the sentence classification task of vanilla finetuning into a masked token prediction task. We finetune the multilingual pretrained language model (MPLM) on source language prompts and apply it to target language data. We use \method with the two MPLMs mBERT and XML-R, and evaluate its efficacy on three different types of multilingual classification tasks in natural language understanding -- multi-class sentiment classification, binary paraphrase identification, and multi-class natural language inference. Our experiments show that \method outperforms vanilla finetuning with both mBERT and XML-R on all three tasks. We further discovered that the performance improvement of \method is generally more obvious in few-shot scenarios. Additionally, we demonstrate that the similarity of the source and target language and the size of the target language pretraining data significantly correlate with the cross-lingual transfer performance of \method, especially on a big dataset with a variety of test languages.

\section*{Limitations}
This study presents the \method pipeline, which aims to enhance zero-shot cross-lingual transfer performance. Our approach was evaluated on various multilingual datasets and showed improved performance. However, due to the limitations of the datasets, only a few languages could be evaluated, thus making it difficult to draw a typological conclusion for all languages. Besides, our exploration in using the prompt-based learning method for cross-lingual language understanding is restricted to single-sentence and sentence pair classifications. As future work, our investigation should be extended to more types of language understanding tasks, such as sequence labelling tasks, e.g. slot detection, named entity recognition, etc. %Besides, as a result of the nature of the prompt architecture for finetuning, only single-head text classification tasks could be included, which prevents the evaluation on other types of tasks (e.g. sequence/slot tagging tasks). %\red{What other types of tasks do you have in mind? Regression? Isn't the 5-sar classification of Amazon already close to a regression task?}

\section*{Ethics Statement}
This research was conducted in accordance with the ACM Code of Ethics. 
All the datasets that we use are publicly available. 
We report only aggregated results in the main paper.
We have not intended or do not intend to share any Personally Identifiable Data with this paper.

\section*{Acknowledgements}
We thank the anonymous reviewers for their efforts and helpful advice. This work was partly supported by Munich Center for Machine Learning (MCML) and China Scholarship Council (CSC).

\bibliography{anthology,custom}
\bibliographystyle{acl_natbib}
%\newpage
\appendix
\section{Appendix}
\label{sec:appendix}

\subsection{Training Details}
\seclabel{training_details}

During training, we used the same hyperparameters for Vanilla and \method to keep the variables consistent for comparison. The chosen hyperparameters for both full-shot training and few-shot training are documented in \tabref{hyperparams}. To avoid random effects on training, we trained each experiment with 5 different random seeds $\{10, 42, 421, 510, 1218\}$ and take the average results.

\begin{table}[h]
  \centering
  \footnotesize
  \renewcommand\arraystretch{1.3}
    \setlength\tabcolsep{5pt}
  \begin{tabular}{lcc}
    \toprule
    Hyperparameter & Full & Few-shot\\
    \midrule
    \textsc{epochs} &    5 &    50\\
    \textsc{learning\_rate}   & 1e-5 &    1e-5\\
    \textsc{batch\_size} &8&1 \\
    \textsc{gradient\_accumulation\_steps} &4&2\\
    \textsc{max\_seq\_length}  &128&128\\
    \textsc{early\_stopping\_patience} & - & 3 \\

    \bottomrule
  \end{tabular}
\caption{Hyperparameters}
\tablabel{hyperparams}
\end{table}

\subsection{Dataset Statistics}
In \tabref{size} we show a basic statistic view of the Amazon Review \citep{keung-etal-2020-multilingual} , PAWS-X \citep{zhang2019paws} and XNLI \citep{williams-etal-2018-broad} datasets. We use the original train-dev-test split from the datasets. For training and validation we use the English train and dev dataset, and for test we use the test sets of all languages. The test data size for each target language is the same in all tasks.

\begin{table}[h]
  \centering
  \renewcommand\arraystretch{1.5}

  \small
  \begin{tabular}{lcccc}
    \toprule
\multirow{2}*{Task}   &  \multicolumn{3}{c}{Size} &  \multirow{2}*{\#Labels}   \\
\cmidrule(){2-4}
       & | Train |  & | Dev |  & | Test | &\\
\midrule
Amazon & $200\ 000$ & $5\ 000$ & $5\ 000$ & 5\\
PAWS-X & $49\ 401$  & $2\ 000$ & $2\ 000$ & 2\\
XNLI   & $392\ 702$ & $2\ 490$ & $5\ 010$ &3 \\

    \bottomrule
  \end{tabular}
\caption{Overview of the three datasets. Train and dev data size refers to the number of samples for English. Test data size refers to the number of samples for each target language.}
\tablabel{size}
\end{table}

\subsection{Reproducibility}

The code for data processing and model training is available at the following Github repository: \url{https://github.com/boleima/ProFiT}.

\subsection{Detailed Results}
\seclabel{detailed_results}
We present the detailed results of few-shot training performance of Vanilla and \method for all three tasks in \tabref{fewshot_amazon} (Amazon Review), \tabref{fewshot_pawsx} (PAWS-X) and \tabref{fewshot_xnli} (XNLI), as well as the T-test results for all tasks in few-shot conditions in \tabref{ttest_fewshot}. %And Table \tabref{feat_per} shows the detailed data used for correlation analysis of language similarity and target data size for all languages and tasks.

\vspace{1em}
% T Test for few shot
\begin{table}[ht]
\renewcommand\arraystretch{1.5}
\setlength\tabcolsep{5pt}
\centering
%\footnotesize
\small
\begin{tabular}{l|cc|cc|cc}
\toprule
\multirow{2}*{Shot} & \multicolumn{2}{c}{Amazon}& \multicolumn{2}{c}{PAWS-X}    &\multicolumn{2}{c}{XNLI}        \\
\cmidrule(){2-7}
     & M  & X & M  & X & M & X  \\
\midrule
1    & 0.001 & 0.001 & 0.50 & 0.56$^*$ & 0.01 & 0.12$^*$ \\
2    & 0.10 & 0.01 & 0.22$^*$ & 0.08$^*$ & 0.89$^*$ & 0.18$^*$ \\
4    & 0.09$^*$ & 0.02 & 0.80$^*$ & 0.10$^*$ & 0.05 & 0.07$^*$ \\
8    & 0.23$^*$ & 0.04 & 0.83$^*$ & 0.04 & 0.86$^*$ & 0.14$^*$ \\
16   & 0.78$^*$ & 0.11$^*$ & 0.30$^*$ & 0.05 & 0.27$^*$ & 0.03 \\
32   & 0.06$^*$ & 0.16$^*$ & 1.00$^*$ & 0.58$^*$ & 0.11$^*$ & 0.01 \\
64   & 0.03 & 0.18$^*$ & 0.02 & 0.80$^*$ & 0.09$^*$ & 0.002 \\
128  & 0.07$^*$ & 0.11$^*$ & 0.15$^*$ & 0.82$^*$ & 0.34$^*$ & 0.01 \\
256  & 0.73$^*$ & 0.21$^*$ & 0.12$^*$ & 0.78$^*$ & 0.07$^*$ & 0.02 \\
512  & 0.86$^*$ & 0.01 & 0.04 & 0.90$^*$ & 0.61$^*$ & 0.004 \\
1028 & 0.003 & 0.31$^*$ & 0.03 & 0.55$^*$ & 0.74$^*$ & 0.03 \\
full & 0.005 & 0.40$^*$ & 0.003 & 0.46$^*$ & 0.005 & 0.44$^*$\\

\bottomrule
\end{tabular}
\caption{T-Test results ($p$) for results of Vanilla and \method in different few-shot conditions. M stands for mBERT and X stands for XLM-R. Insignificant results with a $p$ value $>$ 0.05 are marked with $^*$.}
\tablabel{ttest_fewshot}
\end{table}

%%%%%%%% fewshot on amazon
\begin{table*}

  \centering
  \renewcommand\arraystretch{1.1}
  \setlength\tabcolsep{6pt}
  \footnotesize
  \begin{tabular}{llccccccc}
    \toprule
Shot & Model     & en    & de    & es    & fr    & ja    & zh    & avg.  \\
\midrule
\multirow{4}*{1}    & Vanilla-M & 22.30          & 20.66          & 19.82          & 20.02          & 20.14          & 20.08          & 20.14           \\
     & \method-M  & \textbf{28.52} & \textbf{26.05} & \textbf{26.98} & \textbf{26.18} & \textbf{25.96} & \textbf{25.01} & \textbf{26.04}  \\
\cmidrule(l){2-9}
     & Vanilla-X & 21.98          & 22.15          & 21.69          & 21.79          & 21.42          & 21.52          & 21.71           \\
     & \method-X  & \textbf{37.09} & \textbf{29.86} & \textbf{35.06} & \textbf{36.10} & \textbf{33.13} & \textbf{34.00} & \textbf{33.63}  \\

\midrule
\multirow{4}*{2}    & Vanilla-M & 24.37          & 23.14          & 23.00          & 22.70          & 21.27          & 21.36          & 22.29           \\
     & \method-M  & \textbf{27.63} & \textbf{25.78} & \textbf{26.04} & \textbf{25.05} & \textbf{23.24} & \textbf{23.73}        & \textbf{24.77}  \\
\cmidrule(l){2-9}
     & Vanilla-X & 21.31          & 21.08          & 21.52          & 20.67          & 20.76          & 21.41          & 21.09           \\
     & \method-X  & \textbf{35.63} & \textbf{31.82} & \textbf{33.46} & \textbf{34.40} & \textbf{33.35} & \textbf{32.70} & \textbf{33.14}  \\

\midrule
\multirow{4}*{4}    & Vanilla-M & 27.04          & 24.94          & 23.95          & 23.93          & 23.86          & 22.20          & 23.78           \\
     & \method-M  & \textbf{30.63} & \textbf{26.87} & \textbf{27.67} & \textbf{26.34} & \textbf{25.44} & \textbf{26.05} & \textbf{26.47}         \\
\cmidrule(l){2-9}
     & Vanilla-X & 29.74          & 29.96          & 29.67          & 30.87          & 26.12          & 28.89          & 29.10           \\
     & \method-X  & \textbf{40.23} & \textbf{37.91} & \textbf{38.60} & \textbf{38.75} & \textbf{38.84} & \textbf{37.11} & \textbf{38.24}  \\
\midrule
\multirow{4}*{8}    & Vanilla-M & 29.95          & 26.82          & 26.75          & 26.91          & 24.18          & 25.70          & 26.07           \\
     & \method-M  & \textbf{32.67} & \textbf{29.07} & \textbf{30.20} & \textbf{29.38} & \textbf{26.24} & \textbf{27.12} & \textbf{28.40}  \\
\cmidrule(l){2-9}
     & Vanilla-X & 32.02          & 32.84          & 33.02          & 32.60          & 28.84          & 31.51          & 31.76           \\
     & \method-X  & \textbf{42.23} & \textbf{35.63} & \textbf{40.55} & \textbf{39.79} & \textbf{39.65} & \textbf{38.33} & \textbf{38.79}  \\
\midrule
\multirow{4}*{16}   & Vanilla-M & 33.92          & 30.87          & 32.01          & 30.29          & 28.94          & 28.36          & 30.09           \\
     & \method-M  & \textbf{35.27} & \textbf{31.66} & \textbf{32.10} & \textbf{31.37} & \textbf{29.70} & \textbf{28.58} & \textbf{30.68}  \\
\cmidrule(l){2-9}
     & Vanilla-X & 38.97          & 39.42          & 38.70          & 38.84          & 34.61          & 35.72          & 37.45           \\
     & \method-X  & \textbf{44.78} & \textbf{44.40} & \textbf{43.89} & \textbf{43.55} & \textbf{42.57} & \textbf{41.26} & \textbf{43.13}  \\
\midrule
\multirow{4}*{32}   & Vanilla-M & 36.73          & 31.26          & 31.64          & 31.69          & 28.94          & 29.08          & 30.52           \\
     & \method-M  & \textbf{37.90} & \textbf{33.44} & \textbf{34.68} & \textbf{33.72} & \textbf{31.18} & \textbf{30.77} & \textbf{32.76}  \\
\cmidrule(l){2-9}
     & Vanilla-X & 44.92          & 45.42          & 44.45          & 44.78          & 42.16          & 41.85          & 43.73           \\
     & \method-X  & \textbf{47.51} & \textbf{47.12} & \textbf{46.67} & \textbf{45.78} & \textbf{44.24} & \textbf{42.70} & \textbf{45.30}  \\
\midrule
\multirow{4}*{64}   & Vanilla-M & 39.85          & 33.76          & 35.20          & 34.65          & 30.98          & 29.90          & 32.90           \\
     & \method-M  & \textbf{41.62} & \textbf{36.25} & \textbf{37.84} & \textbf{36.15} & \textbf{32.97} & \textbf{32.56} & \textbf{35.15}  \\
\cmidrule(l){2-9}
     & Vanilla-X & 48.06          & \textbf{48.48} & 46.77          & \textbf{47.34} & 44.01          & 42.05          & 45.73           \\
     & \method-X  & \textbf{49.42} & 48.16          & \textbf{47.99} & 46.93          & \textbf{45.58} & \textbf{44.00} & \textbf{46.53}  \\
\midrule
\multirow{4}*{128}  & Vanilla-M & 43.29          & 35.52          & 37.50          & 36.38          & 32.36          & 31.51          & 34.65           \\
     & \method-M  & \textbf{44.19} & \textbf{38.39} & \textbf{39.84} & \textbf{38.74} & \textbf{34.62} & \textbf{33.71} & \textbf{37.06}  \\
\cmidrule(l){2-9}
     & Vanilla-X & 50.40          & 50.75          & 48.37          & 48.12          & 46.26          & 44.80          & 47.66           \\
     & \method-X  & \textbf{50.75} & \textbf{51.24} & \textbf{49.75} & \textbf{49.22} & \textbf{47.39} & \textbf{45.35} & \textbf{48.59}  \\
\midrule
\multirow{4}*{256}  & Vanilla-M & \textbf{45.64} & 37.15          & 39.23          & 38.20          & \textbf{33.54} & \textbf{32.86} & 36.20           \\
     & \method-M  & 45.39          & \textbf{37.71} & \textbf{39.99} & \textbf{40.31} & 32.55          & 32.82          & \textbf{36.68}  \\
\cmidrule(l){2-9}
     & Vanilla-X & 51.21          & 50.92          & 47.15          & 47.85          & 46.01          & 44.23          & 47.23           \\
     & \method-X  & \textbf{51.40} & \textbf{52.18} & \textbf{50.22} & \textbf{49.81} & \textbf{47.65} & \textbf{45.60} & \textbf{49.09}  \\
\midrule
\multirow{4}*{512}  & Vanilla-M & \textbf{47.66} & \textbf{37.57} & 39.90          & 39.16          & \textbf{33.82} & \textbf{33.64} & 36.82           \\
     & \method-M  & 47.64          & 37.48          & \textbf{40.63} & \textbf{40.99} & 32.76          & 33.40          & \textbf{37.05}  \\
\cmidrule(l){2-9}
     & Vanilla-X & 51.90          & 51.69          & 49.21          & 49.67          & 46.23          & 43.96          & 48.15           \\
     & \method-X  & \textbf{52.94} & \textbf{52.79} & \textbf{50.21} & \textbf{50.06} & \textbf{48.16} & \textbf{45.82} & \textbf{49.41}  \\
\midrule
\multirow{4}*{1024} & Vanilla-M & 49.26          & 38.47          & 41.24          & 39.88          & 33.52          & 33.79          & 37.38           \\
     & \method-M  & \textbf{49.63} & \textbf{41.47} & \textbf{43.54} & \textbf{41.97} & \textbf{36.52} & \textbf{34.54} & \textbf{39.61}  \\
\cmidrule(l){2-9}
     & Vanilla-X & 51.33          & 48.55          & 45.06          & 44.91          & 42.85          & 41.79          & 44.63           \\
     & \method-X  & \textbf{54.55} & \textbf{53.15} & \textbf{51.98} & \textbf{51.18} & \textbf{47.98} & \textbf{46.08} & \textbf{50.07}  \\
\midrule
\multirow{4}*{full} & Vanilla-M & 58.92          & 45.69          & 48.02          & 47.45          & 35.07          & \textbf{38.63} & 42.97           \\
     & \method-M  & \textbf{59.05} & \textbf{46.66} & \textbf{49.30} & \textbf{48.38} & \textbf{37.31} & 38.26          & \textbf{43.98}  \\
\cmidrule(l){2-9}
     & Vanilla-X & 59.61          & \textbf{60.14} & 55.24          & 55.66          & 51.93          & \textbf{49.82} & 54.56           \\
     & \method-X  & \textbf{60.06} & 59.60          & \textbf{55.72} & \textbf{55.89} & \textbf{52.34} & 49.75          & \textbf{54.66}  \\

    \bottomrule
  \end{tabular}
  \caption{Few-shot performance on Amazon}
  \tablabel{fewshot_amazon}
\end{table*}

%%%%%%%% fewshot on pawsx
\begin{table*}

  \centering
  \renewcommand\arraystretch{1.1}
  \setlength\tabcolsep{6pt}
  \footnotesize
  \begin{tabular}{llcccccccc}
    \toprule
Shot & Model     & en    & de    & es    & fr    & ja    & ko    & zh    & avg.  \\
\midrule
\multirow{4}*{1}    & Vanilla-M & \textbf{54.38} & 53.29          & 54.22          & 54.25          & 53.37          & 54.01          & 53.20          & 53.72          \\
     & \method-M & 53.21          & \textbf{54.18} & \textbf{54.44} & \textbf{54.34} & \textbf{55.31} & \textbf{54.35} & \textbf{53.80} & \textbf{54.40} \\
\cmidrule(l){2-10}
     & Vanilla-X & \textbf{51.95} & \textbf{51.75} & \textbf{51.57} & \textbf{51.62} & \textbf{51.95} & \textbf{51.73} & \textbf{51.80} & \textbf{51.74} \\
     & \method-X  & 50.19          & 48.53          & 50.68          & 46.83          & 50.80          & 44.55          & 49.91          & 48.55          \\
\midrule
\multirow{4}*{2}    & Vanilla-M & \textbf{53.54} & \textbf{53.60} & \textbf{53.81} & \textbf{54.18} & \textbf{54.43} & \textbf{54.54} & \textbf{53.77} & \textbf{54.06} \\
     & \method-M  & 52.38          & 53.04          & 53.34          & 53.13          & 54.35          & 53.90          & 51.82          & 53.26          \\
\cmidrule(l){2-10}
     & Vanilla-X & \textbf{54.95} & \textbf{54.73} & \textbf{54.30} & \textbf{54.57} & \textbf{54.25} & \textbf{54.05} & \textbf{54.32} & \textbf{54.37} \\
     & \method-X  & 51.59          & 50.25          & 51.65          & 48.86          & 51.31          & 46.30          & 50.70          & 49.85          \\
\midrule
\multirow{4}*{4}    & Vanilla-M & \textbf{53.93} & \textbf{53.11} & 53.38          & \textbf{53.94} & 53.85          & \textbf{54.28} & \textbf{53.71} & \textbf{53.71} \\
     & \method-M  & 52.40          & 53.07          & \textbf{53.64} & 53.41          & \textbf{54.79} & 53.53          & 51.20          & 53.27          \\
\cmidrule(l){2-10}
     & Vanilla-X & 53.15          & \textbf{54.45} & \textbf{53.99} & \textbf{53.90} & \textbf{53.81} & \textbf{53.79} & \textbf{53.64} & \textbf{53.93} \\
     & \method-X  & \textbf{53.54} & 51.25          & 53.00          & 49.05          & 53.46          & 45.29          & 51.83          & 50.65          \\
\midrule
\multirow{4}*{8}    & Vanilla-M & \textbf{54.30} & 53.50          & \textbf{53.51} & \textbf{54.02} & \textbf{54.03} & \textbf{53.94} & \textbf{54.15} & \textbf{53.86} \\
     & \method-M  & 52.81          & \textbf{54.12} & 53.42          & 53.31          & 53.98          & 53.51          & 51.93          & 53.38          \\
\cmidrule(l){2-10}

     & Vanilla-X & \textbf{54.60} & \textbf{55.13} & \textbf{54.68} & \textbf{54.80} & \textbf{55.46} & \textbf{55.10} & \textbf{55.14} & \textbf{55.05} \\
     & \method-X  & 53.18          & 52.65          & 53.03          & 51.22          & 52.48          & 48.83          & 52.21          & 51.74          \\
\midrule
\multirow{4}*{16}   & Vanilla-M & \textbf{54.08} & 50.86          & 52.04          & 52.66          & 51.77          & 52.27          & 51.23          & 51.81          \\
     & \method-M  & 52.81          & \textbf{53.08} & \textbf{53.80} & \textbf{53.20} & \textbf{53.51} & \textbf{53.95} & \textbf{52.09} & \textbf{53.27} \\
\cmidrule(l){2-10}

     & Vanilla-X & \textbf{54.45} & \textbf{54.84} & \textbf{54.45} & \textbf{54.54} & \textbf{54.96} & \textbf{54.56} & \textbf{54.78} & \textbf{54.69} \\
     & \method-X  & 53.73          & 51.58          & 53.24          & 49.95          & 53.21          & 48.28          & 52.31          & 51.43          \\
\midrule
\multirow{4}*{32}   & Vanilla-M & \textbf{54.03} & 52.94          & 53.48          & \textbf{53.65} & 53.13          & 53.58          & \textbf{53.08} & \textbf{53.31} \\
     & \method-M  & 52.99          & \textbf{52.97} & \textbf{53.75} & 53.14          & \textbf{53.57} & \textbf{54.16} & 51.42          & 53.17          \\
\cmidrule(l){2-10}

     & Vanilla-X & 52.44          & \textbf{53.95} & 52.96          & \textbf{53.21} & 53.46          & \textbf{54.05} & \textbf{53.94} & \textbf{53.60} \\
     & \method-X  & \textbf{53.63} & 51.96          & \textbf{53.44} & 50.51          & \textbf{53.61} & 49.84          & 52.73          & 52.01          \\
\midrule
\multirow{4}*{64}   & Vanilla-M & \textbf{55.44} & \textbf{55.42} & \textbf{55.46} & \textbf{55.97} & \textbf{54.80} & \textbf{55.92} & \textbf{56.41} & \textbf{55.66} \\
     & \method-M  & 53.95          & 54.59          & 54.05          & 54.48          & 54.51          & 54.95          & 52.61          & 54.20          \\
\cmidrule(l){2-10}

     & Vanilla-X & 55.20          & \textbf{55.35} & 54.69          & \textbf{54.95} & \textbf{55.84} & \textbf{55.09} & \textbf{55.39} & \textbf{55.22} \\
     & \method-X  & \textbf{56.60} & 54.95          & \textbf{55.90} & 54.59          & 55.63          & 51.51          & 55.29          & 54.64          \\
\midrule
\multirow{4}*{128}  & Vanilla-M & \textbf{56.63} & \textbf{56.29} & \textbf{56.69} & \textbf{56.43} & 55.31          & 55.70          & \textbf{55.75} & \textbf{56.03} \\
     & \method-M  & 55.54          & 55.76          & 55.28          & 55.26          & \textbf{55.88} & \textbf{55.75} & 55.61          & 55.59          \\
\cmidrule(l){2-10}

     & Vanilla-X & 54.61          & 54.99          & 54.44          & 54.80          & 55.24          & \textbf{55.14} & 54.98          & 54.93          \\
     & \method-X  & \textbf{58.66} & \textbf{56.28} & \textbf{57.95} & \textbf{54.91} & \textbf{56.09} & 52.39          & \textbf{57.35} & \textbf{55.83} \\
\midrule
\multirow{4}*{256}  & Vanilla-M & 58.66          & 56.00          & 56.38          & 56.93          & 55.36          & 55.77          & 55.65          & 56.02          \\
     & \method-M  & \textbf{61.84} & \textbf{60.51} & \textbf{60.65} & \textbf{60.90} & \textbf{58.56} & \textbf{58.70} & \textbf{59.70} & \textbf{59.84} \\
\cmidrule(l){2-10}

     & Vanilla-X & 59.30          & \textbf{58.23} & 58.79          & \textbf{58.54} & 57.18          & \textbf{57.54} & \textbf{57.70} & \textbf{57.99} \\
     & \method-X  & \textbf{59.94} & 57.75          & \textbf{59.58} & 57.86          & \textbf{57.28} & 54.31          & 57.35          & 57.35          \\
\midrule
\multirow{4}*{512}  & Vanilla-M & 64.23          & 59.38          & 60.00          & 60.15          & 56.90          & 56.84          & 56.79          & 58.34          \\
     & \method-M  & \textbf{73.47} & \textbf{69.74} & \textbf{70.23} & \textbf{70.20} & \textbf{63.84} & \textbf{64.56} & \textbf{66.97} & \textbf{67.59} \\
\cmidrule(l){2-10}

     & Vanilla-X & \textbf{77.03} & \textbf{71.28} & 72.09          & \textbf{72.46} & \textbf{63.43} & \textbf{63.79} & 66.53          & \textbf{68.26} \\
     & \method-X  & 76.94          & 71.01          & \textbf{72.29} & 71.24          & 63.19          & 63.28          & \textbf{66.61} & 67.94          \\
\midrule
\multirow{4}*{1024} & Vanilla-M & 74.43          & 68.44          & 69.47          & 70.01          & 61.95          & 61.13          & 64.69          & 65.95          \\
     & \method-M & \textbf{81.06} & \textbf{74.58} & \textbf{76.08} & \textbf{76.15} & \textbf{66.05} & \textbf{66.76} & \textbf{70.64} & \textbf{71.71} \\
\cmidrule(l){2-10}

     & Vanilla-X & 86.33          & \textbf{79.23} & 80.86          & \textbf{80.74} & \textbf{69.25} & \textbf{68.18} & \textbf{73.26} & \textbf{75.25} \\
     & \method-X  & \textbf{87.84} & 78.94          & \textbf{81.53} & 80.58          & 67.68          & 68.01          & 71.85          & 74.76          \\
\midrule
\multirow{4}*{full} & Vanilla-M & 93.85          & 84.94          & 87.11          & 86.55          & 73.39          & 72.44          & 77.01          & 80.24          \\
     & \method-M & \textbf{94.21} & \textbf{86.06} & \textbf{88.17} & \textbf{87.91} & \textbf{75.79} & \textbf{75.82} & \textbf{79.22} & \textbf{82.16} \\
\cmidrule(l){2-10}

     & Vanilla-X & 94.33          & 86.92          & 88.55          & \textbf{89.04} & \textbf{76.07} & 74.71          & 79.75          & 82.51          \\
     & \method-X  & \textbf{94.90} & \textbf{87.06} & \textbf{88.87} & 88.86          & 75.53          & \textbf{75.40} & \textbf{80.63} & \textbf{82.73} 
 \\
    \bottomrule
  \end{tabular}
  \caption{Few-shot performance on PAWS-X}
  \tablabel{fewshot_pawsx}
\end{table*}

%%%%%%%% fewshot on xnli
\begin{table*}

  \centering
  \renewcommand\arraystretch{1.1}
  \setlength\tabcolsep{2.7pt}
  \scriptsize
  \begin{tabular}{llcccccccccccccccc}
    \toprule
Shot & Model     & en             & ar             & bg             & de             & el             & es             & fr             & hi             & ru             & sw             & th             & tr             & ur             & vi             & zh             & avg.           \\
\midrule
\multirow{4}*{1}    & Vanilla-M & 33.58          & 32.97          & 32.97          & 33.46          & 32.70          & 33.33          & 33.43          & 32.44          & 32.93          & 32.85          & 33.12          & 33.05          & 32.96          & 33.00          & 32.99          & 33.02          \\
     & \method-M  & \textbf{37.58} & \textbf{34.93} & \textbf{33.56} & \textbf{35.95} & \textbf{35.02} & \textbf{34.25} & \textbf{36.38} & \textbf{33.93} & \textbf{36.76} & \textbf{34.62} & \textbf{33.83} & \textbf{34.07} & \textbf{34.22} & \textbf{36.43} & \textbf{37.41} & \textbf{35.10} \\
\cmidrule(l){2-18}
     & Vanilla-X & 33.73          & 33.07          & 32.86          & 33.51          & 32.66          & 33.40          & 33.54          & 32.50          & 33.04          & 33.15          & 33.18          & 33.14          & 33.00          & 33.08          & 33.04          & 33.08          \\
     & \method-X  & \textbf{39.26} & \textbf{34.61} & \textbf{34.85} & \textbf{36.28} & \textbf{36.88} & \textbf{33.59} & \textbf{34.92} & \textbf{39.76} & \textbf{34.47} & \textbf{36.53} & \textbf{36.33} & \textbf{36.56} & \textbf{37.03} & \textbf{37.40} & \textbf{36.61} & \textbf{36.13} \\
\midrule
\multirow{4}*{2}    & Vanilla-M & 34.67          & \textbf{34.98} & 36.21          & \textbf{36.15} & \textbf{35.46} & \textbf{36.91} & \textbf{36.42} & 34.34          & 35.42          & \textbf{34.67} & \textbf{34.20} & \textbf{35.40} & 34.04          & 36.18          & 35.61          & \textbf{35.43} \\
     & \method-M  & \textbf{38.38} & 34.85          & \textbf{34.02} & 35.07          & 35.20          & 33.44          & 35.70          & \textbf{35.63} & \textbf{35.65} & 34.50          & 33.78          & 34.35          & \textbf{34.67} & \textbf{36.57} & \textbf{37.12} & 35.04          \\
\cmidrule(l){2-18}
     & Vanilla-X & 34.84          & 34.33          & 35.51          & 35.62          & 34.99          & \textbf{36.25} & 35.86          & 34.14          & 35.09          & 34.39          & 33.95          & 34.87          & 33.76          & 35.55          & 35.02          & 34.95          \\
     & \method-X  & \textbf{39.22} & \textbf{36.54} & \textbf{36.73} & \textbf{38.48} & \textbf{37.83} & 34.21          & \textbf{37.91} & \textbf{38.87} & \textbf{35.56} & \textbf{37.16} & \textbf{38.42} & \textbf{38.01} & \textbf{37.75} & \textbf{38.25} & \textbf{36.98} & \textbf{37.34} \\
\midrule
\multirow{4}*{4}    & Vanilla-M & 37.91          & \textbf{35.47} & \textbf{36.12} & 36.20          & 35.03          & \textbf{36.22} & 36.09          & 34.60          & 35.60          & \textbf{35.01} & \textbf{34.35} & \textbf{35.49} & \textbf{34.49} & 36.28          & 35.74          & 35.48          \\
     & \method-M  & \textbf{38.04} & 35.43          & 34.64          & \textbf{36.67} & \textbf{36.50} & 33.66          & \textbf{36.63} & \textbf{36.07} & \textbf{36.83} & 34.87          & 33.42          & 35.41          & 34.44          & \textbf{37.06} & \textbf{37.07} & \textbf{35.62} \\
\cmidrule(l){2-18}
     & Vanilla-X & 37.55          & 34.31          & 35.08          & 35.11          & 34.09          & \textbf{35.06} & 34.85          & 33.74          & \textbf{34.53} & 34.09          & 33.58          & 34.39          & 33.71          & 35.09          & 34.56          & 34.44          \\
     & \method-X  & \textbf{38.79} & \textbf{36.03} & \textbf{35.23} & \textbf{37.49} & \textbf{37.36} & 33.50          & \textbf{36.54} & \textbf{38.79} & 34.21          & \textbf{37.11} & \textbf{37.79} & \textbf{36.47} & \textbf{37.58} & \textbf{37.96} & \textbf{36.22} & \textbf{36.59} \\
\midrule
\multirow{4}*{8}    & Vanilla-M & \textbf{40.83} & \textbf{37.39} & \textbf{38.56} & \textbf{38.69} & \textbf{37.77} & \textbf{39.25} & \textbf{39.06} & 36.38          & 37.72          & \textbf{37.54} & \textbf{36.46} & \textbf{38.07} & \textbf{36.28} & \textbf{38.22} & 37.76          & \textbf{37.80} \\
     & \method-M  & 38.71          & 36.59          & 35.73          & 37.20          & 37.33          & 34.88          & 38.05          & \textbf{38.22} & \textbf{38.32} & 35.37          & 35.40          & 36.48          & 35.99          & 38.20          & \textbf{38.93} & 36.91          \\
\cmidrule(l){2-18}
     & Vanilla-X & 40.84          & 36.52          & 37.57          & 37.97          & 36.85          & \textbf{38.50} & \textbf{38.35} & 35.70          & 37.00          & 36.77          & 35.57          & 37.33          & 35.57          & 37.56          & 36.95          & 37.01          \\
     & \method-X  & \textbf{41.58} & \textbf{37.81} & \textbf{37.61} & \textbf{39.74} & \textbf{39.06} & 35.07          & 37.65          & \textbf{39.78} & \textbf{37.26} & \textbf{38.64} & \textbf{40.32} & \textbf{38.79} & \textbf{38.65} & \textbf{40.33} & \textbf{38.54} & \textbf{38.52} \\
\midrule
\multirow{4}*{16}   & Vanilla-M & 42.42          & 39.56          & 40.71          & 40.36          & 39.63          & \textbf{41.49} & 41.14          & 37.86          & 39.60          & \textbf{38.27} & 37.35          & 38.77          & 37.44          & 40.76          & 40.25          & 39.51          \\
     & \method-M  & \textbf{44.52} & \textbf{42.10} & \textbf{41.96} & \textbf{40.85} & \textbf{42.18} & 40.63          & \textbf{43.98} & \textbf{41.17} & \textbf{43.10} & 36.50          & \textbf{38.83} & \textbf{41.71} & \textbf{38.95} & \textbf{43.40} & \textbf{43.14} & \textbf{41.32} \\
\cmidrule(l){2-18}
     & Vanilla-X & 42.65          & 39.37          & 40.33          & 40.09          & 39.15          & \textbf{41.12} & 40.73          & 37.72          & 39.44          & 38.02          & 37.34          & 38.63          & 37.19          & 40.73          & 40.01          & 39.28          \\
     & \method-X  & \textbf{49.72} & \textbf{42.15} & \textbf{43.51} & \textbf{47.38} & \textbf{46.22} & 40.19          & \textbf{44.09} & \textbf{45.59} & \textbf{43.14} & \textbf{44.81} & \textbf{46.16} & \textbf{45.39} & \textbf{44.43} & \textbf{47.35} & \textbf{45.69} & \textbf{44.72} \\
\midrule
\multirow{4}*{32}   & Vanilla-M & 46.18          & 40.39          & 41.17          & 41.25          & 40.39          & 42.65          & 41.88          & 38.69          & 40.77          & \textbf{38.29} & 38.47          & 39.62          & 38.82          & 41.18          & 40.89          & 40.32          \\
     & \method-M  & \textbf{49.02} & \textbf{45.64} & \textbf{46.01} & \textbf{44.64} & \textbf{47.57} & \textbf{45.00} & \textbf{48.32} & \textbf{45.06} & \textbf{46.37} & 38.28          & \textbf{43.39} & \textbf{43.68} & \textbf{43.88} & \textbf{47.18} & \textbf{47.78} & \textbf{45.20} \\
\cmidrule(l){2-18}
     & Vanilla-X & 46.11          & 39.69          & 40.44          & 40.57          & 39.81          & 42.05          & 41.28          & 38.30          & 40.25          & 37.71          & 37.99          & 39.05          & 38.17          & 40.27          & 40.00          & 39.68          \\
     & \method-X  & \textbf{52.27} & \textbf{46.87} & \textbf{48.41} & \textbf{49.79} & \textbf{49.12} & \textbf{45.55} & \textbf{48.85} & \textbf{48.42} & \textbf{48.10} & \textbf{45.90} & \textbf{49.20} & \textbf{47.88} & \textbf{46.58} & \textbf{49.84} & \textbf{48.55} & \textbf{48.08} \\
\midrule

\multirow{4}*{64}   & Vanilla-M & 52.10          & 45.26          & 46.64          & 48.10          & 46.32          & 49.44          & 48.57          & 42.71          & 45.45          & 39.13          & 40.24          & 42.19          & 42.41          & 47.23          & 46.91          & 45.04          \\
     & \method-M  & \textbf{55.04} & \textbf{50.28} & \textbf{51.76} & \textbf{52.60} & \textbf{52.90} & \textbf{50.46} & \textbf{53.85} & \textbf{49.57} & \textbf{51.68} & \textbf{42.26} & \textbf{46.38} & \textbf{49.01} & \textbf{48.85} & \textbf{52.89} & \textbf{52.57} & \textbf{50.36} \\
\cmidrule(l){2-18}
     & Vanilla-X & 51.86          & 44.99          & 46.39          & 47.86          & 45.84          & 48.92          & 48.47          & 42.99          & 45.25          & 39.04          & 40.35          & 42.43          & 42.51          & 47.08          & 46.70          & 44.92          \\
     & \method-X  & \textbf{59.35} & \textbf{50.75} & \textbf{53.38} & \textbf{55.47} & \textbf{55.32} & \textbf{50.92} & \textbf{55.71} & \textbf{53.11} & \textbf{52.67} & \textbf{51.31} & \textbf{53.99} & \textbf{52.95} & \textbf{51.30} & \textbf{55.51} & \textbf{54.41} & \textbf{53.34} \\
\midrule

\multirow{4}*{128}  & Vanilla-M & 58.61          & 51.91          & 54.23          & 54.89          & 54.32          & \textbf{56.27} & 55.30          & 49.05          & 52.87          & 43.18          & 46.02          & 49.56          & 48.28          & 54.02          & 54.06          & 51.71          \\
     & \method-M  & \textbf{59.12} & \textbf{53.87} & \textbf{55.09} & \textbf{56.44} & \textbf{55.33} & 55.00          & \textbf{56.09} & \textbf{52.36} & \textbf{54.71} & \textbf{45.25} & \textbf{49.41} & \textbf{52.44} & \textbf{51.35} & \textbf{55.62} & \textbf{55.98} & \textbf{53.50} \\
\cmidrule(l){2-18}
     & Vanilla-X & 58.27          & 51.41          & 53.86          & 54.61          & 53.85          & 55.90          & 54.89          & 48.68          & 52.21          & 42.87          & 46.23          & 49.26          & 47.89          & 53.55          & 53.90          & 51.36          \\
     & \method-X  & \textbf{64.78} & \textbf{56.50} & \textbf{60.23} & \textbf{60.77} & \textbf{60.55} & \textbf{59.51} & \textbf{61.20} & \textbf{57.41} & \textbf{59.13} & \textbf{55.12} & \textbf{58.44} & \textbf{58.15} & \textbf{55.36} & \textbf{60.24} & \textbf{59.68} & \textbf{58.73} \\
\midrule

\multirow{4}*{256}  & Vanilla-M & 61.88          & 53.54          & 56.61          & 57.25          & 56.20          & \textbf{58.77} & 57.91          & 51.31          & 55.45          & 44.97          & 46.97          & 52.75          & 50.07          & 56.51          & 56.76          & 53.94          \\
     & \method-M  & \textbf{62.30} & \textbf{54.82} & \textbf{56.96} & \textbf{57.92} & \textbf{56.48} & 58.69          & \textbf{58.39} & \textbf{53.58} & \textbf{57.09} & \textbf{45.55} & \textbf{49.06} & \textbf{53.64} & \textbf{52.41} & \textbf{57.81} & \textbf{58.06} & \textbf{55.03} \\
\cmidrule(l){2-18}
     & Vanilla-X & 61.68          & 53.30          & 56.19          & 57.01          & 55.91          & 58.47          & 57.74          & 51.13          & 55.22          & 44.86          & 46.68          & 52.77          & 49.79          & 56.24          & 56.33          & 53.69          \\
     & \method-X  & \textbf{66.55} & \textbf{58.08} & \textbf{62.26} & \textbf{62.24} & \textbf{61.23} & \textbf{62.88} & \textbf{63.44} & \textbf{58.56} & \textbf{60.42} & \textbf{54.77} & \textbf{59.95} & \textbf{59.95} & \textbf{56.59} & \textbf{62.28} & \textbf{61.18} & \textbf{60.27} \\
\midrule

\multirow{4}*{512}  & Vanilla-M & 64.94          & 56.75          & 59.66          & 60.73          & 58.53          & \textbf{61.99} & 60.89          & 53.69          & 58.94          & 46.24          & 48.58          & 55.50          & 52.56          & 59.71          & 59.89          & 56.69          \\
     & \method-M  & \textbf{65.39} & \textbf{57.36} & \textbf{60.18} & \textbf{61.03} & \textbf{58.95} & 61.59          & \textbf{61.04} & \textbf{55.07} & \textbf{59.52} & \textbf{47.23} & \textbf{50.48} & \textbf{55.98} & \textbf{54.08} & \textbf{60.25} & \textbf{60.41} & \textbf{57.37} \\
\cmidrule(l){2-18}
     & Vanilla-X & 64.92          & 56.33          & 59.53          & 60.47          & 58.11          & 61.92          & 60.59          & 53.36          & 58.53          & 45.92          & 47.99          & 55.25          & 52.15          & 59.32          & 59.49          & 56.35          \\
     & \method-X  & \textbf{70.13} & \textbf{61.99} & \textbf{66.33} & \textbf{65.47} & \textbf{64.91} & \textbf{67.43} & \textbf{66.72} & \textbf{60.53} & \textbf{64.80} & \textbf{57.27} & \textbf{63.16} & \textbf{63.35} & \textbf{58.78} & \textbf{65.31} & \textbf{64.74} & \textbf{63.63} \\
\midrule

\multirow{4}*{1024} & Vanilla-M & 65.90          & 56.85          & 59.73          & 61.10          & 58.40          & \textbf{62.73} & \textbf{62.07} & 54.57          & 59.38          & 46.46          & 48.46          & 56.19          & \textbf{54.21} & 60.32          & 60.51          & 57.21          \\
     & \method-M  & \textbf{66.77} & \textbf{57.83} & \textbf{59.94} & \textbf{61.53} & \textbf{59.42} & 62.05          & 61.99          & \textbf{55.37} & \textbf{59.54} & \textbf{47.44} & \textbf{49.10} & \textbf{56.40} & 53.91          & \textbf{60.48} & \textbf{60.62} & \textbf{57.54} \\
\cmidrule(l){2-18}
     & Vanilla-X & 65.67          & 56.88          & 59.61          & 60.95          & 57.99          & 62.47          & 61.93          & 54.48          & 59.30          & 46.36          & 48.21          & 56.01          & 54.29          & 60.15          & 60.25          & 57.06          \\
     & \method-X  & \textbf{71.51} & \textbf{63.04} & \textbf{67.62} & \textbf{66.26} & \textbf{66.27} & \textbf{68.64} & \textbf{67.72} & \textbf{62.02} & \textbf{65.86} & \textbf{58.12} & \textbf{64.33} & \textbf{64.41} & \textbf{60.46} & \textbf{66.36} & \textbf{65.50} & \textbf{64.76} \\
\midrule

\multirow{4}*{full} & Vanilla-M & 82.57          & 65.12          & 68.97          & 71.40          & 66.30          & 74.22          & 73.68          & 60.02          & 68.95          & 50.24          & 53.15          & 62.02          & 57.96          & 69.80          & 68.91          & 65.05          \\
     & \method-M  & 82.57          & \textbf{65.55} & \textbf{69.47} & \textbf{71.57} & \textbf{67.43} & \textbf{75.10} & \textbf{74.57} & \textbf{60.57} & \textbf{69.55} & \textbf{51.13} & \textbf{54.58} & \textbf{62.64} & \textbf{58.04} & \textbf{70.74} & \textbf{70.08} & \textbf{65.79} \\
\cmidrule(l){2-18}
     & Vanilla-X & 84.91          & \textbf{71.86} & 77.78          & 76.86          & 75.96          & 79.25          & 78.21          & 69.92          & \textbf{75.79} & \textbf{65.21} & 72.02          & 73.12          & 66.07          & 74.71          & 73.72          & 73.61          \\
     & \method-X  & \textbf{84.97} & 71.81          & \textbf{77.92} & \textbf{77.35} & \textbf{76.11} & \textbf{79.31} & \textbf{78.75} & \textbf{70.10} & 75.43          & 65.13          & \textbf{72.39} & \textbf{73.23} & \textbf{66.95} & \textbf{75.05} & \textbf{73.92} & \textbf{73.82} \\

    \bottomrule
  \end{tabular}
  \caption{Few-shot performance on XNLI}
  \tablabel{fewshot_xnli}
\end{table*}

\clearpage

\end{document}